\documentclass[runningheads]{llncs}

\usepackage{graphicx}
\usepackage{amsfonts}

\begin{document}

\title{Hybrid of DiffStride and Spectral Pooling in Convolutional Neural Networks}

\author{Sulthan Rafif\inst{1},
Mochamad Arfan Ravy Wahyu Pratama\inst{2}, 
Mohammad Faris Azhar\inst{3},
Ahmad Mustafidul Ibad\inst{4}, 
Lailil Muflikhah\inst{5},
\and
Novanto Yudistira\inst{6}} 
\authorrunning{Rafif et al.}
\institute{Faculty of Computer Science, University of Brawijaya\inst{1}\inst{2}\inst{3}\inst{4}\inst{5}\inst{6}
\newline
\email{rsulthan5@gmail.com}\inst{1}, \email{ravyarfan@gmail.com}\inst{2}, \email{varismuvazar@gmail.com}\inst{3},
\newline
\email{amustafidul@gmail.com}\inst{4}, \email{lailil@ub.ac.id}, \inst{5}\email{yudistira@ub.ac.id}\inst{6}}

\maketitle

\begin{abstract}
Stride determines the distance between adjacent filter positions as the filter moves across the input. A fixed stride causes important information contained in the image can not be captured, so that important information is not classified. Therefore, in previous research, the DiffStride Method was applied, namely the Strided Convolution Method with which it can learn its own stride value. Severe Quantization and a constraining lower bound on preserved information are arises with Max Pooling Downsampling Method. Spectral Pooling reduce the constraint lower bound on preserved information by cutting off the representation in the frequency domain. In this research a CNN Model is proposed with the Downsampling Learnable Stride Technique performed by Backpropagation combined with the Spectral Pooling Technique. Diffstride and Spectral Pooling techniques are expected to maintain most of the information contained in the image. In this study, we compare the Hybrid Method, which is a combined implementation of Spectral Pooling and DiffStride against the Baseline Method, which is the DiffStride implementation on ResNet 18. The accuracy result of the DiffStride combination with Spectral Pooling improves over DiffStride which is baseline method by 0.0094. This shows that the Hybrid Method can maintain most of the information by cutting of the representation in the frequency domain and determine the stride of the learning result through Backpropagation.\end{abstract}


\section{Introduction}
The application of downsampling in Convolutions Neural Networks (CNN) generally uses a fixed stride with the aim of reducing the resolution of the image to speed up computation time without reducing the important information contained in the image \cite{Xiang2022}. A fixed stride causes important information contained in the image can not be captured, so that important information is not classified. Therefore, in previous research, the Diffstride Method was applied, namely the Strided Convolution Method with which it can learn its own stride value \cite{Riad2022}. 

The learning process is carried out using the Backpropagation technique \cite{Boue2018}. The learning process using Backpropagation is done by paying attention to each feature in the image. By determining the stride derived from the learning process, every important feature in the image will not be missed. So that important features in the image can be classified.

Severe Quantization and a constraining lower bound on preserved information are arises with Max Pooling Downsampling Method \cite{Rippel2015}. Spectral Pooling reduce the constraint lower bound on preserved information by cutting off the representation in the frequency domain. Thus allowed more information per parameter to be stored.

In this research, a CNN model is proposed with the Downsampling Learnable Stride Technique performed by Backpropagation combined with the Spectral Pooling Technique. Diffstride and Spectral Pooling techniques are expected to maintain most of the information contained in the image by cutting of the representation in the frequency domain and determine the stride of the learning result through Backpropagation. The proposed method is applied to 2D architecture for color image classification.

Based on the training results. The application of the Diffstride combination with Spectral Pooling has an accuracy result of 0.9334. The accuracy result of the Diffstride combination with Spectral Pooling is increasing by 0.0094 from the Diffstride application, namely the baseline method. This shows that the Hybrid Method, which combines Diffstride and Spectral Pooling, can maintain most of the information by cutting of the representation in the frequency domain and determine the stride of the learning result through Backpropagation. To summarize, our contribution can be listed as follows: 

\begin{enumerate}
  \item Hybrid Spectral Pooling and DiffStride give best performance over various experimental setup.
  \item The different positioning of DiffStride and Spectral Pooling affects the performance of the classification.
  \item Spectral Pooling is more effective when placed one level above global average pooling during convolutions, and DiffStride can work best when placed after the first layer of the convolution.
\end{enumerate}

\section{Related Works}

\subsection{Hartley Transform} 
In a previous study conducted by Hao Zhang et al in 2020, the Hartley Transform based on Spectral Pooling was applied for downsampling. Hartley Transform is applied to change the image from the Spatial Domain to the frequency domain. Furthermore, dimension reduction is carried out by selecting a certain frequency subset \cite{Ma2020}. The results of changing the spatial domain to the frequency domain of the image are then processed using spectral pooling. Based on the research conducted, the use of Hartley-based spectral pooling on CNN results in a higher classification accuracy. Then Spectral Pooling can accelerate convergence in the early stages of training.

The Hartley transform can be implemented with the Fourier transform. There is previous research that applies Fourier Transform. Research conducted by Mathieu et al applied Fourier Domain to speed up the training process \cite{Mathieu2014}. This is done by calculating convolution as a product of pointwise in the Fourier Domain and using a feature map that is transformed many times. In this research, the Fast Fourier Transform is applied to change the spatial domain into the frequency domain in the downsampling process

\subsection{Spectral Parameterized}
In the next research, the Fast Fourier Transform was applied to the Spectral Parameters in the downsampling process. In previous research conducted by Rippel et al in 2015 demonstrated the effectiveness of a complex coefficient on the Spectral Parameterized of Convolution Filters \cite{Rippel2015}. Based on this research, the testing for Spectral Pooling showed that it preserves considerably more information for the same number of parameters compared to other pooling strategies like max pooling. It also allows for the selection of any output dimensionality, producing a smooth curve over all frequency truncation choices.

In the next research, a Downsampling Method was applied, applying Spectral Parameters with the Stride Determination Feature through a learning process. In previous research conducted by Rachid Riad in 2022, the DiffStride Method for Downsampling was applied where Strided Convolution can learn its own stride value \cite{Riad2022}. Stride determination done through learning in the Downsampling process can maintain important information contained in the image in the classification process. Based on the research results, Diffstride can provide better performance when compared to the application of Spectral Pooling. Then Diffstride can determine the stride through the learning process, so as to increase the efficiency and accuracy of the model. Resulted in an accuracy of 0.925 for Diffstride and an accuracy of 0.924 for Spectral Pooling. In this research, Spectral Parameters are applied in the Spectral Pooling Method combined with DiffStride in the downsampling process with the aim of improving the performance of the image classification process.

\subsection{Deep Convolutional Networks}
The image classification process can be carried out using deep artificial neural networks. There is previous research that applies deep artificial neural networks for the image recognition process. Research conducted by Simonyan and Zisserman applied Deep Convolutional Network to large-scale image recognition \cite{Simonyan2015}. In this research, it is proposed to increase the depth of the neural network by applying a 3 x 3 convolution filter, which shows significant improvement when compared to the previous architecture. The proposed neural network is applied with 16-19 layers.

In the next research, a Deep Residual Network was applied with the aim of preventing vanishing gradients in the classification process. There is research that applies the Deep Residual Network \cite{He2016}. In that study, an analysis of the implementation of the propagation formulation in residual blocks was conducted, which showed that forward and backward signals can be directly propagated from one block to another, when using identity mappings as skip connections and after-addition activation. Based on the evaluation results, the performance improvement by applying 1001 layer ResNet on CIFAR-10 and CIFAR-100 datasets resulted in 4.62\% error with 200-layer ResNet on ImageNet dataset. ResNet on ImageNet dataset.

There are previous studies that apply Deep Residual to the learning process. Research conducted by He et al. applied a reformulation to the layer applied as a residual function that refers to the input layer \cite{He2016a}. In this study, the application of ImageNet dataset was evaluated on a residual network with a depth of up to 152 layers, which is 8 times deeper than VGG Nets but still has a lower complexity. The ensemble of the proposed residual network obtained an error of 3.57\% on the ImageNet dataset. The test results won the first place in ILSVRC 2015. In this study, 100 and 1000 layers were applied on CIFAR-10 dataset.

In the next research, a Deep Residual Network was applied with 18 Residual Layers. In a previous study conducted by Kaming He et al in 2016 implemented shortcut connections or skip connections which allow information to flow from one residual block to another residual block \cite{Zhang2019}. The ResNet-18 architecture consists of basic blocks called residual blocks. Each Residual Block consists of two convolution layers with the ReLU Activation Function between them.

The contribution from ResNet-18 is that even though the training process uses many layers, the performance generated by the Classification Model does not decrease. This is because information flows from one residual block to another, thus preventing a vanishing gradient from occurring \cite{Hu2018}. In this research, Deep Residual Network is applied in the implementation of Downsampling Techniques, namely Spectral Pooling and DiffStride.

\subsection{Pooling Function}
There are previous studies that developed the ability of the pooling layer to perform feature selection on images. The research conducted by Lee et al. carefully explored approaches to enable pooling to learn and adapt to complex and varied patterns \cite{Lee2016}. In that research, there were two focuses of exploitation: learning the pooling function through the application of two strategies, namely max and average pooling, and learning the pooling function in the form of a structured and self-learning pooling filter. Based on the evaluation results, the pooling generalization process by applying a combination of average and max pooling can improve the performance of the classification model. In this research, a self-learning pooling filter was applied to the DiffStride Method in the downsampling process. In this research, we applied the determination of the optimal learning rate to provide the best performance of the classification model.

\subsection{Batch Normalization}
% Batch normalization: Accelerating deep network training by reducing internal covariate shift
There is previous research that proposes Batch Normalization, namely research conducted by Ioffe and Szegedy \cite{Ioffe2015}. They proposed batch normalization as part of the model architecture by normalizing each training batch. Batch normalization makes it possible to use a high learning rate and does not need to be careful in initialization.

\section{Methods}

\subsection{Hybrid ResNet-18 Architecture}
This study uses the ResNet-18 Architecture Model with the application of DiffStride and Spectral Pooling. DiffStride and Spectral Pooling act as Pooling Layers in the ResNet-18 Architecture. The ResNet-18 architecture is shown in Figure 1.

\begin{figure}[h]
\centering
\includegraphics[width=100pt]{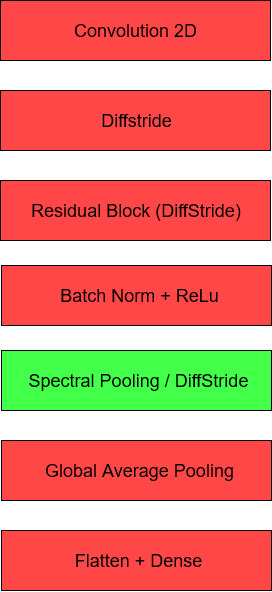}
\caption{Hybrid Spectral Pooling and DiffStride Architectures} 
\end{figure}

Based on the ResNet 18 architecture in Figure 1, this research applies a combination of DiffStride Layer and Spectral Pooling Layer. The Spectral Pooling Layer is placed above the Global Average Pooling Layer. The Spectral Pooling technique performs reduction in the frequency domain by cutting the high frequencies in the image, leaving the low frequencies in the image resulting from the convolution and pooling process. The low frequencies in the image consist of information such as edges and lines and other important features. The image resulting from the convolution and pooling process has a lower dimension. 

A combination of two DiffStride Layers was also applied in this study. The combination of two DiffStride Layers is done by placing the second DiffStride Layer on top of the Global Average Pooling Layer. The second DiffStride Layer plays a role in cutting the high frequency of the image resulting from the convolution and pooling process. The cutting process is based on the determination of the box size through the Backpropagation learning process. 

The determination of the box size through the learning process aims to prevent information from being lost in the convolution and pooling images, so that some information in the image is still stored even though the downsampling process has been carried out. The information that is preserved are features such as edges and lines and other important features. The output result of DiffStride is an image with dimensions that have a small size and contains features that have been selected. So that the next feature extraction process is carried out through a static stride determination. In this study, each residual layer will use DiffStride as the downsampling layer. The application of DiffStride as the downsampling layer for each residual layer is shown in Figure 2.

\begin{figure*}[h]
\centering
\includegraphics[width=350pt]{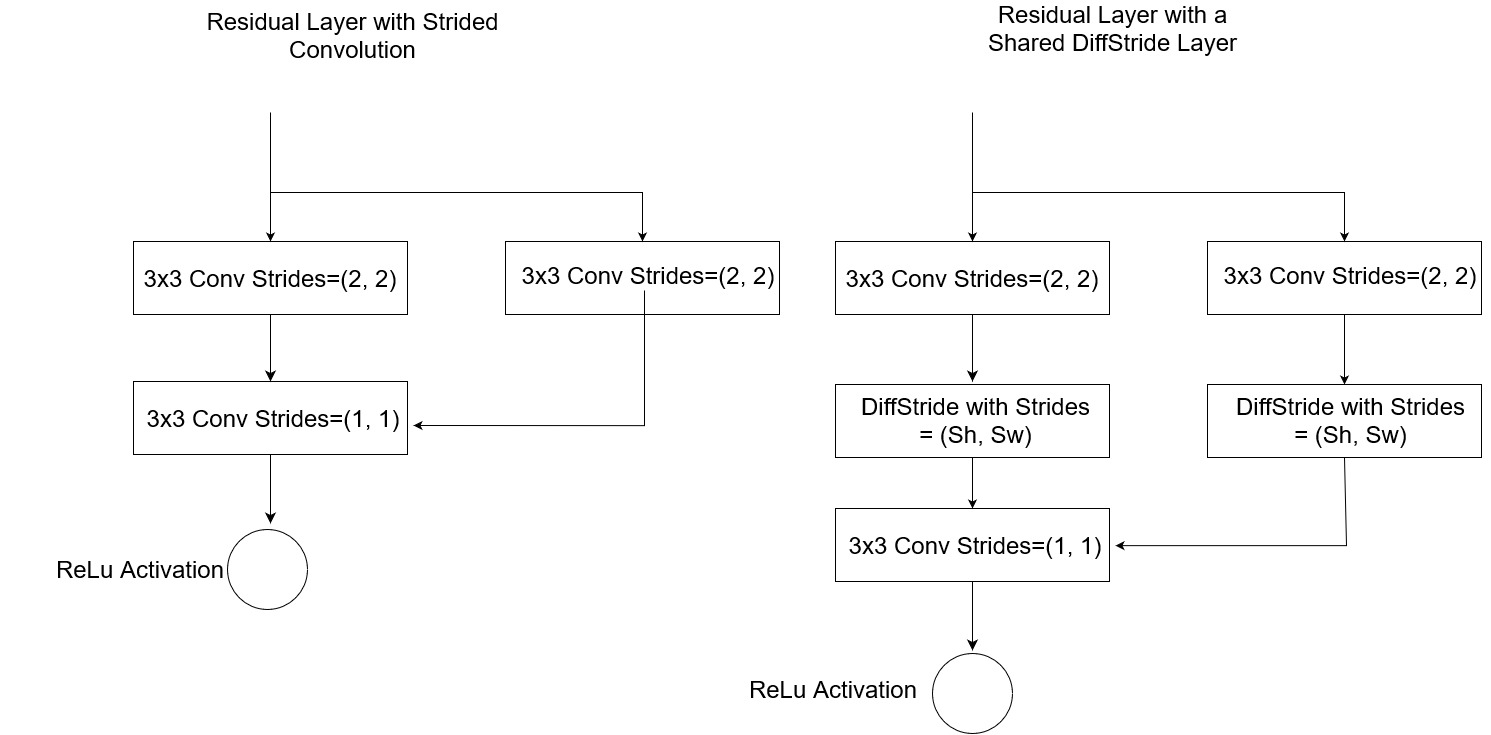}
\caption{Implementation DiffStride in Residual Layer} 
\end{figure*}

Based on Figure 2, ResNet 18 architecture consists of two types of blocks, (1) identity blocks that set the input channel dimension and spatial resolution and (2) shortcut blocks that are used to increase the output channel dimension and reduce the spatial resolution by strided convolution. Based on Figure 2, DiffStride layer is placed on shortcut blocks by replace the Strided Convolution. Shortcut blocks placed on main and residual branches. The application of DiffStride layers in the main and residual branches aims to ensure that each spatial dimension generated by each DiffStride layer is identical and can be summed. The difference between the ResNet 18 Architecture for the Baseline Method and the Hybrid Method, which is a combination of DiffStride and Spectral Pooling or a combination of two DiffStrides is shown in Figure 3.

\begin{figure}[h]
\centering
\includegraphics[width=200pt]{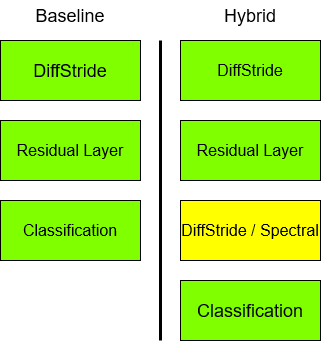}
\caption{Architecture Difference Between Hybrid and Baseline Method} 
\end{figure}

Based on Figure 3, the difference between the ResNet 18 Architecture for the Baseline Method and the Hybrid Method is that the second Spectral Pooling Layer or DiffStride Layer is placed after the Residual Layer and before the Layer for classification, the Dense Layer. DiffStride is applied to implement stride determination based on the learning results to get more important information in the image. Meanwhile, Spectral Pooling aims to cutting of the representation in the frequency domain to store more information.

\subsection{Dataset}
The CIFAR-10 dataset consists of 60,000 color image data and consists of ten classes. With a division of 50,000 training data and 10,000 testing data. The CIFAR-100 dataset has the same number of divisions for training and testing data but consists of one hundred classes. Both CIFAR-10 and CIFAR-100 have 32x32 color images.

The CIFAR dataset is easy to solve, so it can be used as the basis of the process of knowing how to develop and evaluate using convolutional deep learning neural networks for image classification built from scratch. The CIFAR dataset also allows researchers to try many methods or algorithms to evaluate their performance \cite{Ho-Phuoc2018}. In this study, the CIFAR dataset was divided according to the official data, namely 50,000 for training data, and 10,000 for validation data.

\subsection{Spectral Pooling}
Spectral Pooling is a dimensional reduction technique in frequency representation. In this case it is applied to the image signal frequency. This technique makes it possible to use a smaller number of parameters when using other pooling techniques while retaining a lot of information in the image.
Spectral Pooling allows dimensionality reduction by cutting off the representation in the frequency domain. Thus allowing more information per parameter to be stored. Algorithm 1 shows the flow of the Spectral Pooling Layer calculation \newline

\noindent Algorithm 1 Spectral Pooling \newline

\noindent 

\noindent ${\mathbb{R}}$  = \textit{input image }

\noindent $H,\ W$  = \textit{height and width shape}

\noindent $H$   = \textit{Hartley-Transform Function}

\noindent $x\ $  = \textit{input image}

\noindent $\hat{x}$  = \textit{output image}

\noindent $y\ $  = \textit{frequency domain image output}

\noindent $\hat{y}$  = \textit{cropped frequency domain image output} \newline

\noindent 

\noindent Input   : Map $x\mathrm{=}\ {\mathbb{R}}^{H\ \times \ W}$ , output size $h\times w\ $ \newline
\noindent Output : Pooled map $\hat{x}\ \in \ {\mathbb{R}}^{H\ \times \ W}$

\begin{enumerate}
\item Convert input image into frequency domain, with the formula: \newline  $y\ \leftarrow \ H$($x$) 
\item Crop the image based on high frequency, with the formula: \newline  $\hat{y}\ \leftarrow CropSpectrum(y,\ h\ x\ w)$ 
\item Convert frequency domain image into spatial domain, with the formula: \newline  $\hat{x}$ $\ \leftarrow \ H$($\hat{y}$) 
\end{enumerate}

\noindent 

\noindent 

The process of Spectral Pooling consists of three stages. The first step is to insert a spatial domain image with dimensions H x W and convert the image from the spatial domain to the frequency domain using the Discrete Fourier Transform Technique with the Hartley Transform Function \cite{Ma2020}. In the second stage, the cropping process is carried out on the frequency domain image with the aim of separating the low frequencies in the image. In the third stage, the cropped image is then returned to the spatial domain using the inverse of the Hartley Transform Function.

\subsection{Diffstride}
DiffStride performs a learning process to determine the box size using Backpropagation. The learning process using Backpropagation is done by paying attention to each feature in the image. By determining the stride derived from the learning process, every important feature in the image will not be missed. So that important features in the image can be classified. The DiffStride downsampling process is a modification of the Spectral Pooling Downsampling Technique with stride determination done based on learning results. Algorithm 2 shows the flow of the DiffStride Layer calculation. \newline

\noindent Algorithm 2 DiffStride Layer \newline

\noindent 

\noindent ${\mathbb{R}}$   = \textit{input image }

\noindent \textit{R   }= smoothness factor

\noindent $H,\ W$   = \textit{input shapes}

\noindent $F$    = \textit{fourier-transform function}

\noindent $x\ $   = \textit{input image}

\noindent $\hat{x}$   = \textit{output image}

\noindent $y\ $   = \textit{frequency domain image output}

\noindent $\hat{y}$   = \textit{cropped frequency domain image output}

\noindent \textit{S          }  = \textit{strides parameters}

\noindent $h,\ w$   = \textit{spatial coordinate of the stride}

\noindent $\tilde{x}$   = \textit{down sampled output}

\noindent $o$\textit{   }= \textit{element-wise product}

\noindent \textit{Crop   }= \textit{cropping function}

\noindent \textit{mask   }=\textit{ output mask}

\noindent \textit{cropped}  = \textit{output crop} \newline

\noindent Input   : Input $x\mathrm{=}\ {\mathbb{R}}^{H\ \times \ W}$, Strides $S=\left(S_h,\ S_w\right)\ \in \left[\mathrm{1,\ }H\right)\times \ \left[\mathrm{1,\ }W\right),$

smoothness factor R.

\noindent Output : Down sampled output $\tilde{x}\ \in \ {\mathbb{R}}^{\left\lfloor \frac{H}{S_h}\ +\ 2\ \times R\right\rceil \times \ \left\lfloor \frac{W}{S_w}\ +\ 2\ \times R\right\rceil }\ $  \newline

\begin{enumerate}
\item Convert the input image into the Frequency Domain, with the formula \newline $y\ \leftarrow \ F$($x$)   
\item Construct the Mask, with the formula \newline $mask\ \leftarrow W(S_h,\ S_w,\ H,\ W,\ R)$    
\item Apply the mask to filter the high frequency, with the formula \newline y$\ masked\ \leftarrow y\ o\ mask$     
\item Crop the tensor with the mask, with the formula \newline y$\ cropped\ \leftarrow Crop(ymasked,\ sg(mask))$  
\item Convert frequency domain image into spatial domain, with the formula \newline $\tilde{x}\ \leftarrow F^{-1}\ (ycropped)$ 
\end{enumerate}

\noindent 

The DiffStride process consists of five stages. In the first stage, the input image that was originally in the spatial domain is first converted into the frequency domain using the Discrete Fourier Transform Technique by implementing the Fast Fourier Transform Algorithm. Then in the second and third stages, the high-frequency cutting process is carried out on the image with a mask formed using the input shape, strides, and smoothness factor parameters. Furthermore, in the fourth stage, the cropping process is carried out using the mask that has been formed. The cropping results in the frequency domain image are then returned to the spatial domain in the fifth stage.

Unlike Spectral Pooling, the size of the box used to perform the cropping process on the frequency domain image is determined through a learning process. The learning process is done through Backpropagation. The parameters of the box size determination process during learning are the input image size, smoothness factor and step size.

\subsection{Learning Rate Configuration}
Learning Rate is a hyperparameter that affects the effectiveness of conducting the training process on Deep Neural Networks. Determining a good Learning Rate value can influence the training process to find out the right combination of learning rates to achieve convergence \cite{Wu2019}. In this study, we experimented with various learning rate combinations to find out the fastest learning rate combination in converging the training process. Based on the results of learning rate trials, the best learning rate combination is obtained that can achieve convergent results quickly during the training process. The best learning rate combinations used in this study are : (0.1, 0.01, 0.001, 0.0001).

\subsection{Hybrid Diffstride Spectral Pooling}
In this research, a combination of Diffstride with Spectral Pooling is applied to retain important information in the image by cutting the representation in the frequency domain and determining the stride through the learning process. Based on Figure 1, the DiffStride Layer is placed after the 2D Convolution and in the Residual Layer with the aim of reducing high-dimensional images and maintaining most of the information in the image. And the Spectral Pooling Layer is placed before Global Average Pooling with the aim of reducing low-dimensional images that are the result of the convolution and pooling process.

\section{Results and Discussions}
Table 1 shows the training results of applying the entire ResNet-18 architecture using CIFAR-10 Dataset with number of epochs = 200 with learning rate = (0.1, 0.01, 0.001, 0.0001) with various combinations of stride values.

\begin{table}[h]
\caption{CIFAR-10 Accuracy Results}
\centering
\fontsize{7pt}{7pt}\selectfont
\begin{tabular}{cccc}
\hline
Stride Values & DiffStride & DiffStride-Spectral Pool & DiffStride-DiffStride \\
\hline
1, 1, 2, 2, 2 & 0.925 & \textbf{0.9328} & 0.9277  \\
1, 1, 2, 2, 3 & 0.928 & \textbf{0.9341} & 0.9295  \\
1, 1, 1, 3, 1 & 0.924 & \textbf{0.9354} & 0.9231  \\
1, 1, 3, 1, 3 & 0.924 & \textbf{0.9335} & 0.9276  \\
1, 1, 3, 1, 2 & 0.923 & 0.9317 & \textbf{0.9322}  \\
1, 1, 3, 2, 3 & 0.923 & \textbf{0.9329} & 0.9291  \\
Mean Accuracy & 0.924 & \textbf{0.9334} & 0.9282  \\
\hline
\end{tabular}
\end{table}

Based on Table 1, the resulting average val categorical accuracy for the DiffStride Baseline Method is 0.924. And the achievement of convergent results was achieved at the 200th epoch for the application of learning rate = (0.1, 0.01, 0.001, 0.0001) in the DiffStride method, the baseline method. This is evidenced in the graph in Figure 4.

\begin{figure*}[h]
\includegraphics[width=350pt]{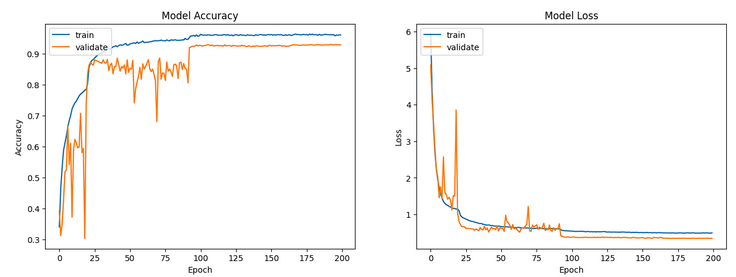}
\caption{Accuracy and Loss for DiffStride (Baseline Method)} 
\end{figure*}

Based on Figure 4, the accuracy of the train and validation datasets for the Diffstride Method (Baseline) starts to be linear at epoch 100 to epoch 200. Based on Table 1, the resulting average val categorical accuracy for the Hybrid Method, which is the combination of Spectral Pooling and DiffStride method is 0.9334. As well as achieving convergent results achieved in the 200 th epoch for the application of learning rate = (0.1, 0.01, 0.001, 0.0001) in the Hybrid Method, a combination of Spectral Pooling and DiffStride. This is evidenced in the graph in Figure 5.

\begin{figure*}[h]
\includegraphics[width=350pt]{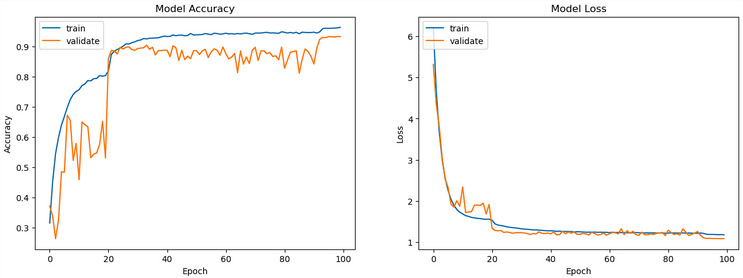}
\caption{Accuracy and Loss for DiffStride Spectral Pooling (Hybrid Method)} 
\end{figure*}

Based on Figure 5, the accuracy and loss of the train and validation datasets for DiffStride Spectral Pooling (Hybrid Method) starts to be linear at epoch 100 to epoch 200. Based on Table 1, the resulting average Val Categorical Accuracy value for the DiffStride Two Combination Method is 0.9282. The value of val categorical accuracy resulting from the two DiffStride methods is lower when compared to the value of val categorical accuracy resulting from the combination of spectral pooling with a average value of val categorical accuracy of 0.9334. As well as achieving convergent results achieved in the 200 th epoch for the application of learning rate = (0.1, 0.01, 0.001, 0.0001) in the Two Combination DiffStride method. This is evidenced in the graph in Figure 6. 

\begin{figure*}[h]
\includegraphics[width=350pt]{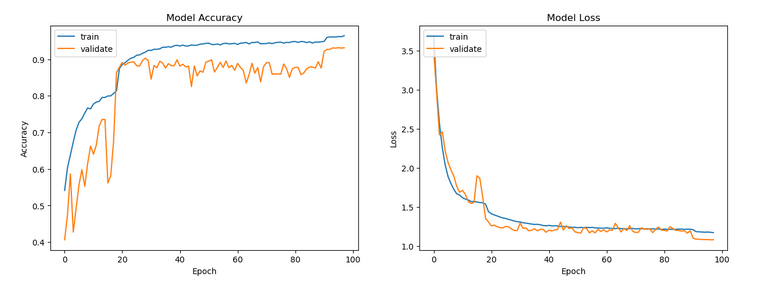}
\caption{Accuracy and Loss for Two DiffStride Combination} 
\end{figure*}

Based on Figure 6, the accuracy and the loss of the train and validation datasets for the DiffStride Two Combination Method starts to be linear at epoch 100 to epoch 200. The Loss value for Training and Validation remains the same after the 100th epoch. Accuracy values for Training and Validation Data also remain after the 100th epoch.

Based on the Table 1, it shows that the DiffStride Downsampling Technique can cope with localized input images, i.e. input images that do not have many dimensional changes. Spectral Pooling Downsampling Technique can overcome the output of image processing results that have made many dimensional changes through convolution and pooling processes. This is because the output of the image that has gone through the convolution and pooling process has a low dimensional size so that the cropping process with the box size determined through the learning process on DiffStride is less efficient. Table 2 shows the training results of applying the Hybrid Method and Baseline Method on the ResNet-18 architecture using CIFAR-100 Dataset with number of epochs = 200 with learning rate = (0.1, 0.01, 0.001, 0.0001) with various combinations of stride values.

\begin{table}[h]
\caption{CIFAR-100 Accuracy Results}
\centering
\begin{tabular}{cccc}
\hline
Stride Values & DiffStride & DiffStride-Spectral Pool \\
\hline
1, 1, 2, 2, 2 & 0.737 & \textbf{0.7446} \\
1, 1, 2, 2, 3 & \textbf{0.737} & 0.7129 \\
1, 1, 1, 3, 1 & 0.703 & \textbf{0.7416} \\
1, 1, 3, 1, 3 & 0.694 & \textbf{0.7416} \\
1, 1, 3, 1, 2 & 0.699 & \textbf{0.7445} \\
1, 1, 3, 2, 3 & 0.666 & \textbf{0.7441} \\
Mean Accuracy & 0.706 & \textbf{0.7382} \\
\hline
\end{tabular}
\end{table}

Based on Table 1 and Table 2, the average of validation categorical accuracy in the Hybrid Method is 0.7382 for CIFAR-100 and 0.9334 for CIFAR-10. The value of validation categorical accuracy is greater than 0.0094 for CIFAR-10 dataset and greater than 0.0322. The value of  validation categorical accuracy is greater than 0.0322 for CIFAR-100 dataset.

\section{Conclusion and Future Works}
Based on the training results, the application of the DiffStride combination with Spectral Pooling has an accuracy result of 0.9334 on CIFAR-10 and 0.7382 on CIFAR-100. The accuracy result of the DiffStride combination with Spectral Pooling is increasing by 0.0094 from the Diffstride application, namely the baseline method for CIFAR-10 Dataset and increasing by 0.0322 for CIFAR-100 Dataset. And the application of the two DiffStride combined method has an accuracy result of 0.9282 on CIFAR-10, where the value is lower by 0.0052 than the combination of Spectral Pooling and DiffStride on the CIFAR-10 Dataset. Spectral Pooling Downsampling Technique can overcome the output of image processing results that have made many dimensional changes through convolution and pooling processes. This shows that the Hybrid Method, which combines Diffstride and Spectral Pooling, can maintain most of the information by cutting of the representation in the frequency domain. However, based on the results of the evaluation, the increase in accuracy is still relatively low. Further research will carry out the process of adding the Method to increase the accuracy of the model.

\end{document}